\documentclass[oribibl,orivec]{llncs}
\usepackage{amsmath2000}
\usepackage{array}
\usepackage{qtree}
\usepackage[numbers,sectionbib,sort&compress]{natbib}
\spnewtheorem{algorithm}{Algorithm}{\bfseries}{\rmfamily}
\spnewtheorem*{proofsketch}{Proof sketch}{\itshape}{\rmfamily}

\newcommand{\liff}{\leftrightarrow}
\DeclareMathSymbol{?}{\mathopen}{operators}{"3F}
\DeclareMathSymbol{\approx}{\mathbin}{symbols}{"19}
\DeclareMathOperator{\Ans}{\textsc{ans}}

\let\bibname\refname

\providecommand{\leanTAP}{\mbox{{\sf lean}\smash{$T^{\!\!\textstyle A}\!\!P$}}}

\newcommand{\tablo}[2]{\ensuremath{\genfrac{}{}{}{0}{#1}{#2}}}
\newcommand{\stablo}[3]{\ensuremath{\genfrac{}{}{}{0}%
    {#1}{#2\;\smash{\vrule height 2.1ex depth .6ex}\;#3}}}

\begin{document}
\title{Question Answering: From Partitions to Prolog%
\thanks{We would like to thank Patrick Blackburn,
Paul Dekker, Jeroen Groenendijk, Maarten
Marx, Stuart Shieber, and the anonymous referees
for their useful comments and discussions.
The 13th European Summer School in Logic, Language and Information, the
13th Amsterdam Colloquium, and
Stanford University's Center for the Study of Language and Information
provided stimulating environments that led to this
collaboration.  The second author is supported by the United States
National Science Foundation under Grant \mbox{IRI-9712068}.}}
\author{Balder ten Cate\inst{1} \and Chung-chieh Shan\inst{2}}
\institute{%
    Institute for Logic, Language and Computation, Universiteit van Amsterdam\\
    Nieuwe Doelenstraat 15, 1012 CP, Amsterdam, The Netherlands\\
    \email{b.ten.cate@hum.uva.nl}
\and
    Division of Applied and Engineering Sciences, Harvard University\\
    33 Oxford Street, Cambridge MA 02138, USA\\
    \email{ccshan@post.harvard.edu}}

\maketitle

\begin{abstract}
We implement Groenendijk and Stokhof's partition semantics of questions in
a simple question answering algorithm.  The algorithm is sound, complete,
and based on tableau theorem proving.  The algorithm relies on a syntactic
characterization of answerhood: Any answer to a question is equivalent to
some formula built up only from instances of the question.  We prove this
characterization by translating the logic of interrogation to classical
predicate logic and applying Craig's interpolation theorem.
\end{abstract}

\section{The Partition Theory of Questions}
\label{s:partition}

An elegant account of the semantics of natural language
questions from a logical and mathematical perspective is the one provided
by \citet{groenendijk-studies}.  According to
them, a question denotes a partition of
a logical space of possibilities. In this section, we give a brief summary
of this influential theory, using a notation slightly different from
\citeauthor{groenendijk-logic}'s presentation \citep{groenendijk-logic}.

A question is essentially a first order formula, possibly with free
variables. We will denote a question by
$?\phi$, where $\phi$ is a first order formula.
(We will also denote a set of questions by $?\Phi$, where $\Phi$
is a set of first order formulas.)
An answer is also a first order formula, but one that stands in a
certain \emph{answerhood} relation with respect to the question,
to be spelled out in~Sect.\,\ref{s:syntactic}.  For example,
the statement ``Everyone is going to the party'' ($\forall x Px$)
will turn out to be an answer to the question ``Who is going to the
party?'' ($?Px$).

% TODO: [Optionally:] On page 1, mention in passing that we skip the part
% of the story of how to compositionally obtain the correct meaning
% from the interrogative NL sentence, jumping straight to partitions.

We assume that equality is in the language, so one can ask questions
such as ``Who is John?'' ($?x\approx j$).
We also assume
that, for every function symbol---including constants---it is indicated
whether it is interpreted rigidly or not.
Intuitively, for a function symbol to be rigid means that its denotation
is known. For example, under the notion of answerhood that we will
introduce in~Sect.\,\ref{s:syntactic}, it is only appropriate to answer
``Who is going to the party?'' ($?Px$) with ``John is going to the party''
($Pj$) if it is known who ``John'' is---in other words, if $j$ is rigid.
Also, for ``Who is John?'' ($?x\approx j$) to be a non-trivial question, ``John''
must have a non-rigid interpretation.

Questions are interpreted relative to first order modal structures
with constant domain. That is, a model is of the form $(W,D,I)$, where
$W$ is a set of worlds, $D$ is a domain of entities, and $I$ is an
interpretation function assigning extensions to the predicates and
function symbols, relative to each world.
Furthermore, we only consider models that give rigid function
symbols the same extension in every world.
Relative to such a model $M=(W,D,I)$, a question~$?\phi$ expresses
a partition of~$W$, in other words an equivalence relation over~$W$:
\begin{equation} \label{e:partition}
[?\phi]_M = \{\, (w,v)\in W^2 \mid \forall g \colon
  M,w,g\models\phi \Leftrightarrow M,v,g\models\phi \,\}
\enspace.
\end{equation}
Roughly speaking, two worlds are equivalent if one cannot tell
them apart by asking the question~$?\phi$.
In general, any set of questions~$?\Phi$
also expresses a partition of~$W$, namely the intersection of the
partitions expressed by its elements:
\begin{equation}
\begin{split}
[?\Phi]_M &= \textstyle\bigcap\nolimits_{\phi\in\Phi} [?\phi]_M \\
          &= \{\, (w,v)\in W^2 \mid \forall \phi\in\Phi\colon \forall g\colon
             M,w,g\models\phi \Leftrightarrow M,v,g\models\phi \,\}
\enspace.
\end{split}
\end{equation}
Entailment between questions is defined as a \emph{refinement} relation
among partitions (i.e.,\ equivalence relations): An equivalence relation~$A$
is a subset of another equivalence relation~$B$ if every equivalence class
of~$A$ is contained in a class of~$B$.
\begin{equation}
?\Phi\models ?\psi \quad\text{iff}\quad
\forall M \colon [?\Phi]_M \subseteq [?\psi]_M
\enspace.
\end{equation}
A more fine-grained notion of entailment is as follows
\citep{groenendijk-logic}. Let $\chi$ be a first order formula
with no free variables, and let $M\models\chi$ mean that
$M,w\models\chi$ for all~$w$.
\begin{equation}
\label{e:entailment}
?\Phi\models_\chi ?\psi \quad\text{iff}\quad \forall M\colon
M\models\chi \Rightarrow
[?\Phi]_M \subseteq [?\psi]_M
\enspace.
\end{equation}
Pronunciation: The questions $?\Phi$ entail the question $?\psi$
\emph{in the context of $\chi$} (or, \emph{given $\chi$}).
The context~$\chi$ is intended to capture assertions in the common ground:
If it is commonly known that everyone who got invited to the party is
going, and vice versa ($\forall x (Ix \liff Px)$), then the questions ``Who
got invited?'' ($?Ix$) and ``Who is going?'' ($?Px$) entail each other.

\subsection{Translation to First Order Logic}

\citet{groenendijk-studies} do not provide an inference system for the above
entailment relation, but we can use the following entailment-preserving
translation procedure to ordinary first order logic.

The intuition behind the translation is simple: One question
entails another iff the former distinguishes between more worlds
than the latter does.  In other words, one question entails another iff
every pair of worlds considered equivalent by the first question is also
considered equivalent by the second.  For instance, the question ``Who is
going to the party?'' ($?Px$) entails the question ``Is John going to the
\pagebreak[3]
party?'' ($?Pj$). After all, if exactly the same people are going to two
parties ($\forall x(Px\liff P'x)$), then either John is going to both
parties, or he is going to neither of them ($Pj\liff P'j$).

For any first order formula $\phi$, let $\phi^*$ be the result of
priming all occurrences of non-rigid non-logical symbols. Formally, define
\begin{align}
(Pt_1\ldots t_n)^* &= P't_1^*\ldots t_n^* &
\smash[b]{\raisebox{-.5\baselineskip}{$f(t_1,\ldots,t_n)^*$}} &
\smash[b]{\raisebox{-.5\baselineskip}{${}=\begin{cases}
        f(t_1^*,\ldots,t_n^*) & \text{if $f$ is rigid} \\
        f'(t_1^*,\ldots,t_n^*) & \text{otherwise}
      \end{cases}$}} \notag\\
(s\approx t)^* &= (s^* \approx t^*) &
\notag\\
(\phi\land\psi)^* &= \phi^*\land\psi^* &
x^* &= x \\
(\neg\phi)^* &= \neg(\phi^*) &
\top^* &= \top \notag\\
(\exists x \phi)^* &= \exists x (\phi^*) &
\bot^* &= \bot \enspace.\notag
\end{align}
Furthermore, for any question $?\phi$, let $?\phi^\#$ be
the first order formula
$%\begin{equation}
    \forall\vec{x}(\phi\liff\phi^*)
$%\end{equation}
, where $\vec{x}$ are the free variables of $\phi$.  For any
set of questions $?\Phi$, let $?\Phi^\#$ be the set of first order
formulas $\{\,?\phi^\#\mid\phi\in\Phi\,\}$.
Now, we can reduce entailment between questions to ordinary first
order entailment, as follows.

\begin{theorem}\label{thm:translation}
The following entailments are equivalent.
\begin{enumerate}
\item $?\Phi\models_\chi ?\psi$
\item $?\Phi^\#,\chi,\chi^* \models ?\psi^\#$
\end{enumerate}
\end{theorem}

\begin{proof}
{[$\Rightarrow$]}\@ By contraposition. Suppose
$?\Phi^\#,\chi,\chi^*
  \not\models ?\psi^\#$. Then there is a first order model
  $M=(D,I)$ that verifies the formulas $?\Phi^\#,\chi,\chi^*$
  but not~$?\psi^\#$.

Now consider the first order modal structure $N=(\{w,v\},D,I')$, where
for all non-logical symbols~$\alpha$,
we let $I'_w(\alpha)=I(\alpha)$ and $I'_v(\alpha)=I(\alpha')$.
By construction, $(w,v)\in [?\Phi]_N$
and $(w,v)\not\in [?\psi]_N$. Furthermore, $N\models\chi$.
Therefore, $?\Phi\not\models_\chi ?\psi$.

{[$\Leftarrow$]}\@ Again by contraposition. Suppose
$?\Phi\not\models_\chi ?\psi$.
Then there is a first order modal structure $M=(W,D,I)$ with $w,v\in W$
such that $(w,v)\in [?\Phi]_M$ yet $(w,v)\not\in [?\psi]_M$. Furthermore,
$M,w\models\chi$ and $M,v\models\chi$.

Now consider the first order model $N=(D,I')$, where
for all non-logical symbols~$\alpha$,
we let $I'(\alpha)=I_w(\alpha)$ and $I'(\alpha')=I_v(\alpha)$.
By construction, $N$ verifies the formulas $?\Phi^\#,\chi,\chi^*$
but not $?\psi^\#$. Therefore,
$?\Phi^\#,\chi,\chi^* \not\models ?\psi^\#$.\qed
\end{proof}

By this theorem, we can determine whether an entailment between
questions holds, by translating the formulas involved to ordinary
predicate logic. For example, suppose that $a$ is a rigid
constant, and consider the supposed entailment
\begin{equation}
    ?Px\models ?Pa \enspace.
\end{equation}
In order to determine whether the entailment is valid, we need only
translate the two formulas to ordinary first order logic, giving
\begin{equation}
    \forall x (Px\leftrightarrow P'x) \models Pa\leftrightarrow P'a \enspace.
\end{equation}
As can be easily seen, this is indeed valid. On the other hand, if $a$
were non-rigid then the entailment would not be valid.
As a second example, consider
\begin{equation}
    ?Px\models_{\forall x (Px\liff Qx)} ?Qa \enspace.
\end{equation}
Given that $a$ is rigid, this is again valid, as we can see from the
translation
\begin{equation}
    \forall x (Px\leftrightarrow P'x), \forall x (Px\liff Qx),
    \forall x (P'x\liff Q'x) \models Qa\liff Q'a \enspace.
\end{equation}
%stopzone

\subsection{Varying Domains}
\label{s:varying}

The partition theory as formulated in the previous section makes some
natural predictions regarding question entailment.  For example, the
question ``Who is going to the party?'' ($?Px$) entails the question ``Is
everyone going to the party?'' ($?\forall x Px$).  Intuitively, whenever
one knows (completely) who is going to the party, one also knows whether
everyone is going to the party. As a matter of fact, much of the
motivation for the partition theory of questions comes from its
natural account of entailment relations between sentences
embedding such questions.\footnote{Cf.~\citet{nelken-calculus} for
one of the few competing theories in this respect.} 

Unfortunately, the theory also makes some
counterintuitive predictions. For example,
the question $?j\approx b$ (``Is John the same 
person as Bill?'', where $j$ and $b$ are interpreted rigidly) is entailed by
every question, including the trivial question $?\top$. 
Likewise, $?\exists x\exists y\neg(x\approx y)$
(``Does there exist more than one entity?'') is entailed by
every question.%
\footnote{In Sect.\,\ref{s:syntactic}, we will see exactly which 
such counterintuitive predictions are made by the theory. For
now, the reader can use Theorem~\ref{thm:translation} to 
check that the theory indeed makes these predictions.}
For this reason, an alternative semantics has been discussed, in which 
the questions are
interpreted with respect to first order modal structures with
\emph{varying} domains \citep{groenendijk-questions}.
That is, each world $w$ is associated with its own
domain of entities $D_w$. This semantics is more general, since it
allows more models: Every constant
domain model is also a varying domain model, so fewer
entailments between questions are valid for varying domains. 
In particular, $?\exists x\exists y\neg(x\approx y)$ is no longer entailed by 
every question. Also,
$?Px$ no longer entails $?\neg Px$, though
the entailment $?Px,?x\approx x \models ?\neg Px$
remains valid.

It is not entirely trivial to generalize the partition semantics of
questions to varying domains, since it is unclear in~\eqref{e:partition}
which assignments~$g$ should be quantified over.
One generalization, which \Citet{groenendijk-questions} seem to suggest but
do not state, is to define
\begin{equation}
\label{e:vdpartition}
\begin{split}
[?\phi]_M = \{\, &(w,v)\in W^2 \mid \forall g \colon \\
 &(g\in D_w^\phi\ \&\ M,w,g\models\phi) \Leftrightarrow
  (g\in D_v^\phi\ \&\ M,v,g\models\phi) \,\}
   \enspace,
\end{split}
\end{equation}
%stopzone
where $D_w^\phi$ denotes the set of assignment functions that map all free
variables in~$\phi$ to entities in~$D_w$.
Question entailment, as defined in~\eqref{e:entailment}, remains the same.

Various generalizations of the partition theory to varying domains,
including that in~\eqref{e:vdpartition}, can be straightforwardly
reduced to the constant domain version by introducing an
\emph{existence predicate}~$E$.  In the case of~\eqref{e:vdpartition}, one
would \emph{relativize} all quantifiers---that is, replace $\exists x \phi$
by $\exists x (Ex\land\phi)$ and $\forall x \phi$ by $\forall
x (Ex\to\phi)$---and consider questions of the form $?(Ex_1\land
\ldots\land Ex_n\land\phi)$, where $x_1$,~\dots,~$x_n$ are
the free variables of~$\phi$.  Theorem~\ref{thm:translation} can then be
applied after the reduction.

In the rest of this paper, we will restrict ourselves to constant 
domain models.

\section{A Syntactic Characterization of Answerhood}
\label{s:syntactic}

Groenendijk and Stokhof's entailment relation between questions, as
discussed in the previous section, allows us to define a notion
of answerhood.
\begin{definition}[Answerhood]
\label{def:answerhood}
Let $?\phi$ be a question and
$\psi$ a first order formula without free variables. We say that $\psi$ is an
\emph{answer} to $?\phi$ if $?\phi\models?\psi$.
\end{definition}
According to this notion (also termed \emph{licensing} by 
\citet{groenendijk-logic}), ``Everyone is going to the party''
($\forall x Px$) is an answer to ``Who is going to the
party?'' ($?Px$), because $?Px \models ?\forall x Px$.
(We will generalize to sets of questions in Sect.\,\ref{s:multiple}.)

Note that, under this definition, any contradiction or tautology counts as
an answer to any question.  Groenendijk and Stokhof
\citep{groenendijk-studies,groenendijk-logic} define a stricter notion of
\emph{pertinence}, which excludes these two trivial cases by formalizing
Grice's Maxims of Quality and Quantity, respectively.  In this paper,
however, we will stick to the simpler criterion of answerhood as defined
above, which corresponds to Grice's Maxim of Relation.

We now have a semantic notion of answerhood, telling us what
counts as an answer to a question. However, for practical purposes, it
is useful to have also a \emph{syntactic} characterization of this
notion. Can one give a simple syntactic property that is a necessary
and sufficient condition for answerhood? As we will see in a minute,
one can.

First, let us look at a partial result discussed by
\citet{groenendijk-studies} and
\citet{kager-questions}. Define rigidity of
terms and formula instances as follows.

\begin{definition}[Rigidity]
\label{def:rigidity}
A term is \emph{rigid} if it is composed of variables and rigid
  function symbols.
A formula~$\phi$ is a \emph{rigid instance} of another formula~$\psi$
  if $\phi$ can be obtained from
  $\psi$ by uniformly substituting rigid terms for variables.
An identity statement $s\approx t$ is \emph{rigid} if the terms $s$ and $t$
  are rigid.
\end{definition}
For example, if $c$ is a rigid constant and $f$ is a rigid function symbol,
then rigid instances of $Px$ include $Pc$, $Px$, $Pf(c)$, and $Pf(x)$. The
identity statement $c\approx x$ is also rigid. Notice that rigid instances
are not necessarily rigid: If $c$ is rigid, then $Rcd$ is a 
rigid instance of $Rxd$ even if the constant $d$ is not rigid. 

\citet{groenendijk-studies} and \citet{kager-questions} observed that rigid
instances of a question constitute answers to that question. By a
simple inductive argument, one can generalize this a bit.

\begin{definition}[Development]
\label{def:development}
A formula~$\psi$ is a \emph{development} of another
  formula~$\phi$ (written ``$\phi\leq \psi$'') if $\psi$ is built up from
rigid instances of $\phi$ and rigid identity statements using
boolean connectives and quantifiers, or if $\psi$ is $\top$ or
$\bot$.
\end{definition}
For example, if $c$ and~$d$ are rigid constants, then $Pc\land Pd$ and
$\exists x (Px\land \neg(x\approx c))$ are both developments of $Px$.

\begin{theorem}\label{thm:gsk}If $\phi\leq \psi$ then $?\phi\models ?\psi$.
\end{theorem}

\begin{proof}
By induction on the size of $\phi$. \qed
\end{proof}
Using the translation procedure in Sect.\,\ref{s:partition},
together with Craig's interpolation theorem for first order logic,
we can prove the converse as well.

\begin{theorem}\label{thm:syntactic}
Let $\vec{y}$ be the free variables of some formula~$\psi$.
If $?\phi\models_\chi ?\psi$, then there exists some
formula~$\vartheta$
with no free variables beside $\vec{y}$
such that $\phi\leq \vartheta$ and $\chi \models \forall\vec{y}(\psi
\leftrightarrow\vartheta)$.
\end{theorem}

\begin{proof}
First, we will prove the special case where $\phi$ is an atomic formula,
say~$P\vec{x}$.  Suppose that $?P\vec{x} \models_\chi ?\psi(\vec{y})$.
Then, by Theorem~\ref{thm:translation},
\begin{equation}
\forall\vec{x}(P\vec{x}\leftrightarrow P'\vec{x}),\chi,\chi^* \models
\forall\vec{y}(\psi(\vec{y})\leftrightarrow\psi^*(\vec{y}))
\enspace.
\end{equation}
As a fact of first order logic, we can replace the universally
quantified variables in the consequent by some freshly chosen constants $\vec{c}$.
This results in
\begin{equation}
\forall\vec{x}(P\vec{x}\leftrightarrow P'\vec{x}),\chi,\chi^* \models
\psi(\vec{c})\leftrightarrow\psi^*(\vec{c})
\enspace,
\end{equation}
and, from this,
\begin{equation}
\label{e:interpolatee}
\forall\vec{x}(P\vec{x}\leftrightarrow P'\vec{x}), \chi^*, \psi^*(\vec{c})
\models \chi\to\psi(\vec{c})
\enspace.
\end{equation}
By Craig's interpolation theorem for first order logic, we can
construct an interpolant $\vartheta(\vec{c})$ such that
\begin{gather}
\label{e:interpolation-1}
\forall\vec{x}(P\vec{x}\leftrightarrow P'\vec{x}), \chi^*, \psi^*(\vec{c})
    \models \vartheta(\vec{c}) \enspace,\\
\label{e:interpolation-2}
\vartheta(\vec{c}) \models \chi\to\psi(\vec{c}) \enspace,
\end{gather}
and the only non-logical symbols in $\vartheta(\vec{c})$ are those
occurring on both sides of~\eqref{e:interpolatee}.
From the way the translation procedure $(\cdot)^*$ is set up, it follows
that the only non-logical symbols that $\chi^*$ and~$\psi^*$ on the one hand,
and $\chi$ and~$\psi$ on the other hand, have in common, are rigid
function symbols.  Thus, $\vartheta(\vec{c})$ contains no
non-logical symbols beside $P$, $\vec{c}$, and rigid function symbols.

Removing primes uniformly from all predicate and function symbols in
 \eqref{e:interpolation-1}, we get
 $\chi\models\psi(\vec{c})\to\vartheta(\vec{c})$.
From \eqref{e:interpolation-2}, we get the converse:
 $\chi\models\vartheta(\vec{c})\to\psi(\vec{c})$. Together, this gives us
\begin{equation}
    \chi\models\psi(\vec{c})\leftrightarrow \vartheta(\vec{c})
    \enspace.
\end{equation}
Since the constants $\vec{c}$ do not occur in $\chi$,
we can replace them by universally quantified variables.
This results in
\begin{equation}
    \chi\models \forall \vec{y} (\psi(\vec{y})\leftrightarrow
    \vartheta(\vec{y}))
    \enspace.
\end{equation}
Furthermore, $\vartheta(\vec{y})$ contains no non-logical symbols
beside $P$ and rigid function symbols. From this, it
follows that $\vartheta(\vec{y})$ is a development
of $P\vec{x}$.

As for the general case, suppose $?\phi(\vec{x})\models_\chi ?\psi$.
Choose a fresh predicate symbol~$P$ with the same arity as the number
of free variables of $\phi$. Then it follows that
$?P\vec{x} \models_{\chi\land\forall\vec{x}(P\vec{x}\liff
  \phi(\vec{x}))} ?\psi$. Apply the above strategy to obtain a development
$\vartheta$ of $?P\vec{x}$ such that
$\chi\land\forall\vec{x}(P\vec{x}\leftrightarrow
\phi(\vec{x}))\models \forall\vec{y}(\vartheta\leftrightarrow \psi)$.
Let $\vartheta'$ be the result of replacing all subformulas in
$\vartheta$ of the form $P\vec{z}$ by $\phi(\vec{z})$. Then
$\vartheta'$ is a development of $\phi$, and
$\chi\models\forall\vec{y}(\psi\leftrightarrow \vartheta')$.\qed
\end{proof}
Thus, the syntactic notion of development corresponds precisely to the
semantic notion of entailment between questions.

Recall that the formula~$\chi$ in Theorem~\ref{thm:syntactic} 
represents an assertion in the common ground.
If no assumptions on the common ground are made (i.e., $\chi=\top$),
then Theorem~\ref{thm:syntactic} reduces to the following syntactic
characterization of answerhood.

\begin{corollary}\label{cor:answerhood}
$\psi$ is an answer to $?\phi$ iff $\psi$ is
  equivalent to a development of $\phi$.
\end{corollary}

This syntactic characterization is useful for several purposes. First
of all, it makes possible a thorough investigation of the predictions
\pagebreak[3]
made by Groenendijk and Stokhof's theory of answerhood: It shows
what their semantic theory really amounts to, syntactically speaking.

Second, this result opens the way to practical question answering
algorithms. Now that we have a syntactic characterization of answerhood,
we can address the question answering problem purely in terms of
symbolic manipulation without having to refer to the semantics. This
will be the topic of Sect.\,\ref{s:finding}.

Another advantage is that it becomes possible to compare different
theories of questions and answers. For instance, while at first sight
Prolog and the partition theory of questions seem incomparable, we
will see in Sect.\,\ref{s:prolog} that they are in fact closely related.

\subsection{Languages without Equality}

What if we do not have equality in the language? This is an interesting
question for at least two reasons. First, as we saw in
Sect.\,\ref{s:varying}, a
number of counterintuitive predictions of the partition semantics
involve equality. For example, sentences such as $\neg(j\approx
b)$ (``John is not Bill'', where $j$ and $b$ are interpreted rigidly) and
$\exists x\exists y \neg (x\approx y)$ (``There exist at least two 
entities'') are answers to every question.
Removing equality would prevent these predictions, though then answers like
``Only John is going to the party'' would no longer be expressible.

The second reason why eliminating equality from the language is
interesting is more practical. In
question answering algorithms, it is convenient
not to have to deal with equality, since equality reasoning is very
expensive. From the theorem proving literature,
one can conclude that dealing with equality is not feasible for many
practical applications.

As it turns out, the syntactic characterization result is even simpler
for first order languages without equality. The corresponding notion of
\emph{development of~$\phi$} is simply \emph{a formula built from
rigid instances of~$\phi$ using boolean connectives and
quantifiers, or $\top$ or~$\bot$}. The same proof of Theorem~\ref{thm:syntactic} goes through,
since Craig's interpolation theorem holds regardless of whether
equality is present.

\subsection{Multiple Questions}
\label{s:multiple}

The notion of answerhood that we introduced earlier
(Definition~\ref{def:answerhood}) generalizes trivially to multiple
questions: If $\psi$ is a first order formula without free variables, we
say that $\psi$ is an answer to~$?\Phi$ if $?\Phi \models ?\psi$.  The
notion of development (Definition~\ref{def:development}) also generalizes:
A formula~$\psi$ is a development of~$\Phi$ if $\psi$ is built up from
rigid instances of elements of~$\Phi$ and rigid identity statements using
boolean connectives and quantifiers, or if $\psi$ is $\top$ or~$\bot$.
The proof of Theorem~\ref{thm:syntactic} then generalizes directly to multiple
questions, as does Corollary~\ref{cor:answerhood}:
$\psi$ is an answer to~$?\Phi$ iff $\psi$ is equivalent to
a development of~$\Phi$.

Multiple questions arise naturally in two settings. First, a query like
``Who got invited to the party, and who is going to the party?''
corresponds to a set of questions like $?\Phi=\{?Ix,?Px\}$.
According to the above generalization of answerhood, one answer
to this query is $\forall x (Ix \liff Px)$: ``Everyone invited is going,
and vice versa.''

Second, suppose that, as part of their common ground, the questioner and
the answerer both know the complete true answers to a certain set of
questions~$?\Theta$.  For example, if
it is commonly known who got invited, then $?\Theta$ contains~$?Ix$.
Extend the notion of entailment in~\eqref{e:entailment} to handle
questions as contexts in the following way: If $\chi$ is a first order
formula with no free variables, then
\begin{alignat}{2}
    ?\Phi \models_{\chi,?\Theta} ?\psi
    \quad\text{iff}\quad
&   \forall M = (W,D,I)\colon
\\&
    M\models\chi \text{, }
    [?\Theta]_M = W^2
    \Rightarrow
    [?\Phi]_M \subseteq [?\psi]_M
    \enspace.\notag
\end{alignat}
%stopzone
%
Intuitively, given the common ground~$?\Theta$,
a formula~$\psi$ is an answer to a question~$?\phi$ just in case
the entailment $?\phi \models_{?\Theta} ?\psi$ holds.
Indeed, the entailment
\begin{equation}
    ?Px \models_{?Ix} ?\forall x (Ix \liff Px)
\end{equation}
is valid: If it is commonly known who got invited, then in response to
``Who is going to the party?'' one can answer ``Everyone invited is going,
and vice versa.''  In fact, we have
\begin{equation}
    ?\Phi \models_{\chi,?\Theta} ?\psi
    \quad\text{iff}\quad
    ?\Theta, ?\Phi \models_\chi ?\psi \enspace.
\end{equation}
In particular,
\begin{equation}
    ?\phi \models_{?\Theta} ?\psi
    \quad\text{iff}\quad
    ?\Theta, ?\phi \models ?\psi \enspace.
\end{equation}
This result means that asking a question~$?\phi$ under some common
ground~$?\Theta$ is exactly like asking the set of questions
$?\Theta\cup\{?\phi\}\cup$ under no common ground, in terms of what counts as
an answer.  So multiple questions arise again
\citep{groenendijk-logic}.

We are about to present a question-answering algorithm.  For simplicity, we
will assume that there is only a single question to answer, and that the
context is empty.  The algorithm we will present generalizes
trivially to handle multiple questions at once, and hence
to handle questions in the context.%
\footnote{In passing, we mention that the semantics of multiple questions
  can be reduced to that of single questions, provided that the domain
  contains at least two elements. More specifically, any two questions
  $?\phi_1$ and~$?\phi_2$ are equivalent to the single question
  $?\bigl((x\approx y\land\phi_1)\lor(\neg(x\approx y)\land\phi_2)\bigr)$.
  This reduction is of practical importance to the reader who encounters
  a fairy that promises to answer a single eternal burning question.}

\section{Finding the Answer to a Question}
\label{s:finding}

The general task of question answering is the following.
\begin{quote}\it
Given a finite first order theory~$\Sigma$ and a question~$?\phi$,
find an answer to~$?\phi$ that is entailed by~$\Sigma$
and that is as informative as possible.
\end{quote}
In other words, we wish to find an informative formula~$\psi$, such
that $\Sigma\models\psi$ and $?\phi\models?\psi$.  Here the theory~$\Sigma$
is intended to capture the answerer's private knowledge, and
we measure how \emph{informative} an answer is by its logical strength: One
answer is more informative than another if the former entails the latter.
For example, an answerer who knows that John is going to the party
and that it rains ($\Sigma = \{Pj,R\}$) might reply to ``Who is going to the party?'' ($?\phi = ?Px$) with
the statement ``Someone is going to the party'' ($\psi = \exists x Px$).
\pagebreak[3]
If the constant symbol~$j$ is rigid, then the statement ``John is going to
the party'' ($Pj$) would be an answer and preferred because it is more
informative.

The first question that comes to mind when considering the task of
question answering is: \emph{Is there always an optimal answer?} In other
words, given a finite theory~$\Sigma$ and a question~$?\phi$, is there
always (modulo logical equivalence) a unique most informative answer
to~$?\phi$ entailed by~$\Sigma$? The answer is \emph{no}.

\begin{theorem}\label{thm:no-optimal}
  There is a finite theory $\Sigma$ and a question $?\phi$ such
  that, for every answer to $?\phi$ entailed by $\Sigma$, there is a
  strictly more informative answer to $?\phi$ entailed by $\Sigma$.
\end{theorem}

\begin{proof}
Let $\Sigma$ be the theory
\begin{equation}
\{ \forall xyz (x<y\land y<z\to x<z), \
\forall x \neg (x < x),\  \forall x \exists y (x<y) \}
\end{equation}
(``$<$ is an unbounded strict order''), and
$\psi_n$ ($n\in\bbbn$) be the formula
\begin{equation}
\textstyle
\exists x_1\ldots \exists x_n \bigwedge_{i,j\leq n\text{; }i\neq y} \neg (x_i\approx x_j)
\end{equation}
(``there are
at least $n$ different objects''). Every $\psi_n$ is entailed by
$\Sigma$. Furthermore,
every $\psi_n$ is an answer to the question
$?\top$.  (This follows either via Theorem~\ref{thm:gsk}, or directly from
how question entailment was defined in~\eqref{e:entailment}.)
The optimal answer would have to entail each $\psi_n$ and,
furthermore, contain no non-logical symbols. It follows from the
compactness theorem that there is no such formula. \qed
\end{proof}

Equality is not essential to this counterexample.
Indeed, the same argument goes through if we replace the
equality sign~$\approx$ by a non-logical binary relation $I$,
use $?Ixy$ as the question, and extend the theory $\Sigma$ with
\emph{replacement axioms}~\citep{fitting-first}:
\begin{equation}
\textstyle
\begin{split}
\Sigma^+=\Sigma\;\cup\;\{\ & \forall x Ixx, \
\forall xyzu( Ixz\land Iyu\land x<y\to z<u ), \\
& \forall xyzu( Ixz\land Iyu\land Ixy\to Izu )\ \}
\enspace.
\end{split}
\end{equation}
Moreover, in the absence of equality there is a second problem,
concerning undecidability.
It is undecidable whether the given formula is the most
specific answer to a question entailed by a theory. This
follows from a simple reduction argument: Suppose we do not have
equality in the language, and suppose the theory $\Sigma$
does not contain any rigid function symbols. Then $\Sigma$ is satisfiable
iff $\top$ is the most specific answer to $?\top$ entailed by $\Sigma$.
But as we all know, first order satisfiability is undecidable.

\begin{figure}
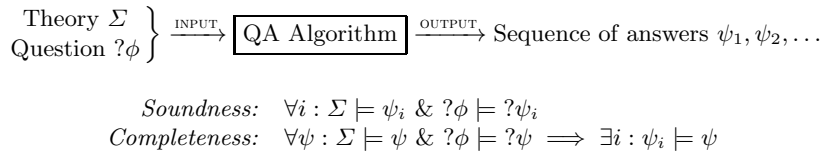

\begin{center}
\begin{tabular}{c}
    $
    \left.
    \begin{tabular}{c}Theory $\Sigma$\\Question $?\phi$\end{tabular}
    \right\}
    \xrightarrow{\textsc{input}}
    \fbox{QA Algorithm}
    \xrightarrow{\textsc{output}}
    \text{Sequence of answers $\psi_1,\psi_2,\ldots$}
    $
\\
\noalign{\bigskip}
    \begin{tabular}{r@{\quad}l}
    \text{\textit{Soundness:}} &
        $\forall i: \Sigma\models\psi_i\ \&\ ?\phi\models ?\psi_i$
    \\
    \text{\textit{Completeness:}} &
        $\forall \psi : \Sigma\models\psi\ \&\ ?\phi\models ?\psi
         \implies \exists i : \psi_i\models\psi$
    \end{tabular}
\end{tabular}
\end{center}
\caption{A question answering algorithm}
\label{f:pipeline}
\end{figure}

Notwithstanding these negative results, it is possible to construct a
sound and complete question answering algorithm. The output of the
algorithm is a sequence of answers that is \emph{cofinal}
in the set of all answers entailed by the theory~$\Sigma$. Cofinality
means that, for any answer that is entailed by $\Sigma$, there is a
more informative answer in the sequence generated by the
algorithm. Formally:
\begin{definition}[Cofinality]
\label{def:cofinality}
A set of formulas $\Phi$ is \emph{cofinal} in another set of
formulas~$\Psi$ if each formula in~$\Psi$ is entailed by some formula
in~$\Phi$.
\end{definition}
We call a question answering algorithm \emph{sound} if it generates
only formulas that are answers to the question and that are entailed
\pagebreak[3]
by the theory. We call the algorithm \emph{complete} if
the sequence it generates is always cofinal in the set
of all answers to the question entailed by the theory.
A question answering algorithm with these properties is
depicted in Fig.\,\ref{f:pipeline}.

To show that there are indeed algorithms satisfying these constraints,
we will now discuss a rather trivial algorithm. Recall that,
to answer a question~$?\phi$ given a theory~$\Sigma$, it suffices to find
developments of~$\phi$ that are entailed by~$\Sigma$.  The search for
answers can be conducted using any theorem proving technique, including
tableaux.  A first stab at a question answering algorithm, then, is to
syntactically enumerate and check all potential answers~$\psi$.

\begin{algorithm}
\label{alg:horrible}
To answer the question~$?\phi$ given the theory~$\Sigma$,
repeat in dovetail fashion for every development~$\psi$ of~$\phi$:
\begin{enumerate}
\item Initialize a tableau with a single branch, consisting of $\Sigma$
      and~$\neg\psi$.
\item Keep applying tableau expansion and closure rules.
\item If the tableau becomes closed, report $\psi$ as an answer.\qed
\end{enumerate}
\end{algorithm}
This algorithm is sound and complete, because tableau-based first order
theorem proving is.  However, it is also horribly inefficient, since it
considers every development of~$\phi$ as a potential answer, without taking
the theory~$\Sigma$ into account.

The challenge, then, is to construct an question answering algorithm
that is sound and complete (like Algorithm~\ref{alg:horrible}) yet
as efficient as possible. In the next
section, we will make a start by providing a more intelligent algorithm.

\subsection{Tableau-Based Question Answering}

We will now introduce a sound and complete question answering
algorithm based on free variable tableaux \citep{fitting-first}.
Without loss of generality, we assume that the question~$?\phi$
consists of a single atomic formula $P\vec{x}$, called the \emph{answer
literal}~\citep{green-application}.  (If that is not so, simply add the
formula $\forall\vec{x}(P\vec{x}\leftrightarrow\phi)$ to the theory, where
$P$ is a new predicate symbol and $\vec{x}$ are the free variables
of~$\phi$.)  We also assume that the theory~$\Sigma$ is Skolemized;
that is, any existential quantifier in~$\Sigma$ has been eliminated with
the help of additional, non-rigid function symbols.
Finally,
we will disregard equality for the
moment, but the algorithm can be extended to deal with
equality as well (we will return to this issue at the end of this section).

Table~\ref{tab:tablorules} summarizes the tableau calculus
for first order logic without equality given by \citet{fitting-first}.
(We omit the existential rule since the theory is Skolemized.)
\pagebreak[3]

\begin{table}
\caption{Tableau expansion and closure rules for theorem proving}
\label{tab:tablorules}
\centering
\begin{tabular}{lc}
Conjunction &
\tablo{\phi\land\psi}{\phi,\psi} \qquad\qquad
\stablo{\neg(\phi\land\psi)}{\neg\phi}{\neg\psi}  \\ \\
Quantification & \tablo{\neg\exists x \phi}{\neg\phi[x/y]} \quad where $y$ is fresh \\ \\
Negation & \tablo{\neg\neg\phi}{\phi}\\ \\
Closure &
\begin{tabular}[t]{l}
  A branch is closed if it contain a formula and its negation.\\
  A tableau is closed if all its branches are closed.\\
  A substitution $\sigma$ is \emph{closing}
    if the tableau is closed after applying~$\sigma$.
\end{tabular}
\end{tabular}
\end{table}
%stopzone
%
\noindent
In the algorithm that we will introduce shortly, besides the usual
tableau expansion rules, one other operation is allowed. At any stage
of tableau construction, one is allowed to create a copy of
the question---renaming its free variables to new variables---and
either add the copy to all branches or add its negation to all
branches. We term this operation the \emph{Add Instance} rule,
shown in Table~\ref{tab:addinstance}.

\begin{table}
\caption{The Add Instance rule for question answering}
\label{tab:addinstance}
\centering
\begin{tabular}{lc}
Add Instance &
    $\overline{     P\vec{y}}$
    \hspace{5em}
    $\overline{\neg P\vec{y}}$
    \vspace{1mm}\\
& \quad
  where $?P\vec{x}$ is the question, and $\vec{y}$ are fresh variables\quad
\end{tabular}
\end{table}

Note that, whereas the rules in Table~\ref{tab:tablorules} are only
  applied to a single branch at a time, the Add Instance rule is always applied
  to the entire tableau. This difference is not essential---completeness
  is not lost if we restrict the application of the Add
  Instance rule to single branches---but it increases efficiency and it
  renders the generated answers more concise.

\begin{figure}
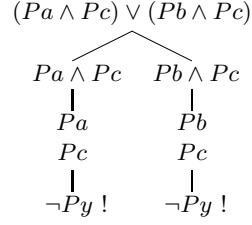

\Tree [.{$(Pa\land Pc)\lor(Pb\land Pc)$}
   [.{\ \ $Pa\land Pc$\ \ } [.{$Pa$\\$Pc$} {\ $\neg Py$\ !} ] ]
   [.{\ \ $Pb\land Pc$\ \ } [.{$Pb$\\$Pc$} {\ $\neg Py$\ !} ] ] ]
\caption{A partially developed tableau}
\label{fig:extablo}
\end{figure}

\begin{example}
A partially developed tableau is given in
Fig.\,\ref{fig:extablo}, where we are given the theory
$\Sigma=\{(Pa\land Pc)\lor(Pb\land Pc)\}$ and wish to answer the
question~$?Px$. Note the use of the Add Instance rule,
indicated by~``!''.
Also note that $y:=c$ is a closing substitution: After
applying it, the tableau is closed.
\end{example}

Every closing substitution generates an answer, in the following way.
Suppose we face a partially developed tableau~$T$, and it can be closed
by a substitution~$\sigma$. Let~$\Phi$ be the (finite) set of all
instances of the question that have been added so far using the Add
Instance rule.  Then $\neg\bigwedge\Phi^\sigma$ is an appropriate answer.
In fact, we can even do a little better and give a slightly more
informative answer, by only considering those added instances that
participate in closure.
\begin{definition}[Closure]
Let~$T$ be a tableau, $\sigma$ be a substitution over~$T$, and $\kappa$ be
a set of formula occurrences in~$T$.  The pair~$(\sigma,\kappa)$ is
a \emph{closure} if $\kappa^\sigma$ contains, on each branch, some formula
and its negation.
\end{definition}
For every closing substitution~$\sigma$, there is at least one
closure~$(\sigma,\kappa)$.  Let~$\Phi_\kappa$ consist of exactly those formulas
in~$\Phi$ that appear in~$\kappa$.  With $\Ans(\sigma,\kappa)$ we will
denote the result of universally quantifying all free variables
in~$\neg\bigwedge(\Phi_\kappa)^\sigma$.

\begin{example}
Consider again the tableau in Fig.\,\ref{fig:extablo}.  Let~$\sigma$ be the
substitution $y:=c$, and let~$\kappa$ contain the occurrences of $Pc$ and
$\neg Py$ on both branches.  Then the pair~$(\sigma,\kappa)$ is a closure.
It generates the answer $\Ans(\sigma,\kappa)=\neg\neg Pc$.
\end{example}

When generating answers from closures, we will only be interested in the
\emph{most general} closures.  One
closure~$(\sigma_1,\kappa_1)$ is more general than
another~$(\sigma_2,\kappa_2)$ if $\sigma_1$ is
more general than~$\sigma_2$ and $\kappa_1$ is a subset of~$\kappa_2$.
Closures that are more general do as little as possible besides closing
the tableau, and so generate answers that are more informative.
When there is more than one most general closure, we compute all the
corresponding answers and take their conjunction.

One remaining difficulty concerns rigid function symbols.  According
to our syntactic characterization of answerhood, non-rigid constants
and function symbols are not allowed to occur in generated
answers.  For instance, suppose the theory is $\{Pc\}$, where $c$ is
a non-rigid constant. To the question $?Px$, we must answer $\exists
x Px$ rather than $Pc$.  Moreover, Skolem functions created during
tableau expansion are also considered non-rigid and disallowed in
answers. For instance, given the theory $\{\exists x Px\}$ and the
question $?Px$, we must answer $\exists x Px$, rather than ``$Ps$,
where $s$ is the Skolem constant such that $Ps$''.  Our algorithm
achieves this by applying Chadha's \emph{unskolemization} procedure
\citep{chadha-applications}.

The complete question answering algorithm is as follows.

\begin{algorithm}\label{alg:notsobad}To answer the question $?P\vec{x}$
given the theory $\Sigma$, start by initializing a tableau for $\Sigma$.
Next, do one of the following, repeatedly, \emph{ad infinitum}.
\begin{enumerate}
\item Apply a tableaux expansion rule (Table~\ref{tab:tablorules}).
\item Apply the Add Instance rule (Table~\ref{tab:addinstance}).
\item Take the conjunction $\psi_0=\bigwedge\Ans(\sigma,\kappa)$ over all
      most general closures~$(\sigma,\kappa)$.  Unskolemize~$\psi_0$ to
      remove any non-rigid or Skolem function symbols, and report the
      result as the next answer.\qed
\end{enumerate}
\end{algorithm}

\begin{theorem}
Algorithm~\ref{alg:notsobad} is sound and complete, provided the
non-deterministic choices are made in a \emph{fair} manner.
\end{theorem}
\begin{proofsketch}
Call a formula~$\psi$ a \emph{pre-answer} if $\psi$ is built up from
instances of the answer literal~$P\vec{x}$---allowing non-rigid and Skolem
function symbols---using boolean connectives and quantifiers, or if
$\psi$ is $\top$ or $\bot$.  We first prove that the algorithm, minus
unskolemization, generates pre-answers in a sound and complete
manner.  Soundness means that the algorithm only generates
(yet-to-be-unskolemized) conjunctions~$\psi_0$ that are pre-answers and
that are entailed by the theory.  Completeness means that the sequence of
conjunctions generated is cofinal in the set of all pre-answers entailed by
the theory.

Soundness: Because $\Ans(\sigma,\kappa)$ is always a pre-answer, so is any
generated conjunction~$\psi_0$.  Moreover, any closure $(\sigma,\kappa)$
for our question-answering tableau gives rise to to a closed
theorem-proving tableau for $\Sigma\cup(\Phi_\kappa)^\sigma$.  Since
tableau theorem proving is sound, the theory~$\Sigma$ must entail
$\Ans(\sigma,\kappa)$.  Hence $\Sigma$ entails~$\psi_0$.

Completeness: Suppose that $\psi$ is a pre-answer entailed by~$\Sigma$.
By completeness of tableau theorem proving, we can find a closed
theorem-proving tableau for $\Sigma\cup\{\neg\psi\}$, and systematically
transform it into a question-answering tableau for $(\Sigma,?P\vec{x})$
that generates a pre-answer~$\psi_0$ entailing~$\psi$.  Given this, it can
be shown that in fact any fair question-answering tableau expansion
procedure will eventually generate a pre-answer entailing~$\psi$.

We now need to relate pre-answers to answers.
The unskolemization procedure is sound; that is, unskolemizing~$\psi_0$
always gives a formula entailed by~$\psi_0$.  Besides, unskolemizing any
pre-answer gives an answer, so Algorithm~\ref{alg:notsobad} is sound.  The
unskolemization procedure is also complete; that is, whenever
a pre-answer~$\psi_0$ entails some answer~$\psi$, the result of
unskolemizing~$\psi_0$ also entails~$\psi$.  Besides, every answer is
a pre-answer, so Algorithm~\ref{alg:notsobad} is complete.\qed
\end{proofsketch}

One way to guarantee fairness in Algorithm~\ref{alg:notsobad} is to
implement it with \emph{depth first iterative deepening}, just as
\citet{beckert-leantap} did in their {\leanTAP} prover.
In fact, we have modified {\leanTAP} to become a lean
question answerer.

The difference between Algorithms \ref{alg:horrible}
and~\ref{alg:notsobad} is that the latter waits until closing the
tableau before deciding which answer to prove to follow from the theory.
In other words, Algorithm~\ref{alg:notsobad}
does not commit to the rigid instances that constitute the
development until they are determined by the closure.
This ability to postpone commitment is exactly the strength of
  free variable tableau calculi as compared to ground tableaux
  \citep{fitting-first}.

This algorithm can be extended to deal with equality in two steps.
First, add the tableau rules necessary for theorem proving with
equality \citep{beckert-adding,fitting-first}. Second, generalize the
Add Instance rule, so that not only instances of the question but also
(in)equalities can be added to the tableau.

\subsection{Prolog as a Special Case}
\label{s:prolog}

A precise connection can be established between Prolog and our algorithm.
Prolog performs question answering in a sense more restrictive than
considered here, because it makes extra assumptions about the
theory~$\Sigma$ and the question~$?\phi$:
\begin{enumerate}
\item The theory~$\Sigma$ is required to be in Skolemized Horn form and
      the question must consist of a single atom, in order to make
      computation feasible.
\item There is no equality predicate in the basic language of Prolog.
\item Prolog assumes that all function symbols and constants are rigid,
      making $Pc$ a potential answer to $?Px$ even if the symbol~$c$
      resulted from Skolemization.
\item Due to its depth-first search strategy, Prolog is complete only for
      some theories, for example theories without
      cycles among predicate symbols.
\end{enumerate}
Subject to all these restrictions, Prolog is an optimal question answering
algorithm: Given a theory and a question, it produces answers optimal in
the sense of the partition theory of questions.%
\footnote{%
One apparent difference between Prolog and our algorithm
is that Prolog is based on resolution, whereas we used tableaux.
However, the two methods are closely related, and Prolog can be
interpreted as a variant of so-called \emph{connection tableaux}
\citep{haehnle-tableaux}.}

As we can see, Prolog is a question answering algorithm that
  makes many extra assumptions.  The present work makes it possible to
  distinguish and identify these assumptions, and eliminate them.
  A broad spectrum of generalizations of Prolog can then be considered.

\section{Conclusion}

This paper makes two contributions. First, we presented a syntactic
characterization of answerhood for the partition semantics of
questions. The applications of this result are mainly internal to the
partition semantics: It explains the meaning of a question in terms of
the form of its answers.

Second, our tableau-based question answering algorithm connects two
important research traditions: question answering systems such as Prolog,
and the formal semantics of natural language questions. We feel that the
link between these two fields of research has been neglected in the
past, and hope to bring them together.

We want to mention three directions of further research (we stress the
third).
\begin{description}
\item[Logical] Some theoretical issues are still to be addressed:
  How can answerhood be characterized syntactically when the semantics
  allows varying domains? Also, is it decidable whether a given answer
  is the optimal answer to a question entailed by a theory? Without
  equality, this problem is undecidable, as we proved in
  Sect.\,\ref{s:finding}. We have yet to find a similar reduction
  argument for the case with equality.

\item[Linguistic] We want to bring our theoretical results
  to bear on the semantics of questions in natural language.
  In particular, our syntactic characterization result
  clarifies the linguistic predictions made by the
  partition semantics~\citep{shan-partition}.

\item[Computational] We want to use this
  work as a unifying framework to compare different approaches to question
  answering.  In particular, we want to investigate a variety of Prolog
  generalizations and determine which assumptions made by Prolog are
  feasible to drop.

  \quad
  Like us, \citet{green-application} and \citet{luckham-extracting} have
  applied theorem proving to question answering, but their criteria for
  answerhood are tied to the syntax of formulas in prenex normal form.
  This difference explains why their algorithms omit unskolemization,
  a step necessary for soundness under the notion of answerhood we adopt
  here.  Also, \citet{bos-first} have devised a simplified version of the
  partition semantics for computational purposes. These efforts seem to fit
  nicely in our picture. 
\end{description}

\bibliographystyle{mcbride}
\bibliography{p2p}

\end{document}